\documentclass[runningheads]{llncs}

\usepackage{eccv}

\usepackage{eccvabbrv}

\usepackage{graphicx}
\usepackage{booktabs}

\usepackage[accsupp]{axessibility}  

\usepackage{hyperref}

\usepackage{orcidlink}

\begin{document}

\title{OrbitNVS: Harnessing Video Diffusion Priors for Novel View Synthesis} 

\titlerunning{OrbitNVS: Harnessing Video Diffusion Priors for Novel View Synthesis}

\author{Jinglin Liang$^{1,3}$ \and
Zijian Zhou$^{2}$ \and
Rui Huang$^{1}$ \and
Shuangping Huang$^{1,4}$\thanks{Corresponding Author} \and
Yichen Gong$^{3}$
}

\authorrunning{J. Liang et al.}

\institute{$^{1}$South China University of Technology,
$^{2}$King's College London, \\
$^{3}$Agentic Labs, XGTech, 
$^{4}$Pazhou Laboratory \\
\email{eeljl@mail.scut.edu.cn, eehsp@scut.edu.cn}}

\maketitle
\begin{center}
\centering
\includegraphics[width=\linewidth]{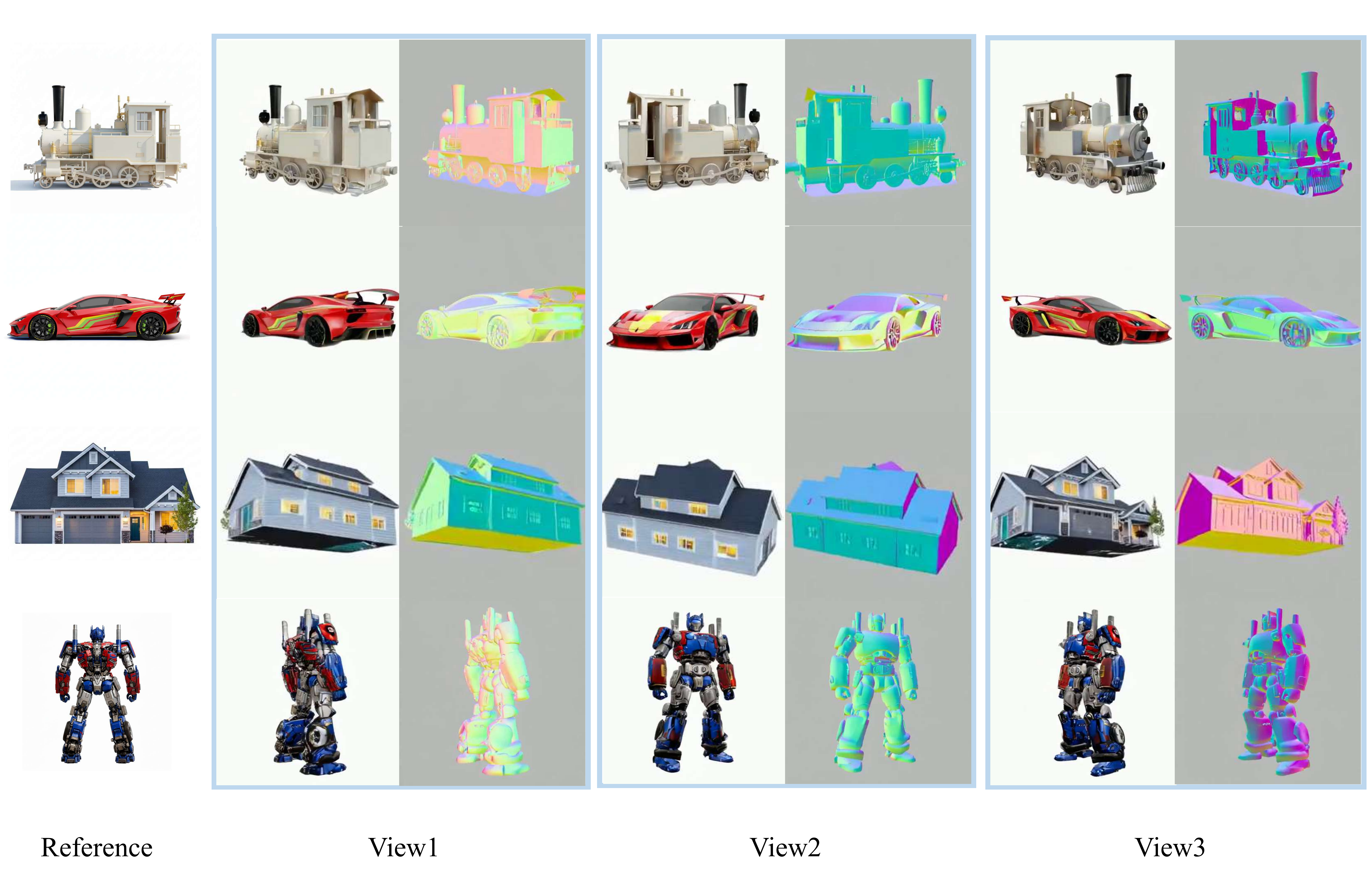}
\captionof{figure}{OrbitNVS leverages the rich visual prior of video generation models to achieve highly consistent NVS with precise camera control. Its strong prior enables impressive reasoning capabilities in the unseen views. For instance, it can infer the detailed front view of a robot from only its back, or deduce the presence of windows on the opposite side of a house from a single frontal image.}
\label{fig:cover}
\end{center}

\begin{abstract}
Novel View Synthesis (NVS) aims to generate unseen views of a 3D object given a limited number of known views.
Existing methods often struggle to synthesize plausible views for unobserved regions, particularly under single-view input, and still face challenges in maintaining geometry- and appearance-consistency.
To address these issues, we propose OrbitNVS, which reformulates NVS as an orbit video generation task.
Through tailored model design and training strategies, we adapt a pre-trained video generation model to the NVS task, leveraging its rich visual priors to achieve high-quality view synthesis.
Specifically, we incorporate camera adapters into the video model to enable accurate camera control.
To enhance two key properties of 3D objects, geometry and appearance, we design a normal map generation branch and use normal map features to guide the synthesis of the target views via attention mechanism, thereby improving geometric consistency.
Moreover, we apply a pixel-space supervision to alleviate blurry appearance caused by spatial compression in the latent space.
Extensive experiments show that OrbitNVS significantly outperforms previous methods on the GSO and OmniObject3D benchmarks, especially in the challenging single-view setting (\eg, +2.9 dB and +2.4 dB PSNR).
\end{abstract}

\section{Introduction}
\label{sec:intro}
Novel View Synthesis (NVS) aims to generate photorealistic unseen views of a object using only a limited set of input views \cite{shi2023zero123++,Liu2023zero1to3}. 
This capability promises to revolutionize a spectrum of applications, powering the creation of 3D assets for film and gaming \cite{zhao2025hunyuan3d}, enabling immersive AR/VR experiences \cite{qu2024nerf}, and advancing robotic perception \cite{bar2025navigation}.

The NVS task is inherently an ill-posed problem. Its core challenge lies in ``imagining'' the appearance of unseen areas based on a limited set of seen views. 
Humans address this challenge by leveraging visual commonsense knowledge, which is acquired through extensive observation of the physical world, to infer occluded regions \cite{liu2025scaling}. 
For instance, when observing the front of an object, humans can infer its back by relying on priors of similar objects, without explicitly constructing a precise 3D model in the mind.
This observation indicates that the NVS is fundamentally a problem of visual commonsense reasoning rather than a geometric reconstruction. 
Given that video data is rich in visual commonsense and is significantly cheaper to acquire than 3D data, which requires expensive expert modeling or specialized scanning equipment, video data and the generative models pre-trained on it present a promising pathway for advancing NVS.

Several existing works have attempted to learn such visual commonsense from large-scale multi-view image data. For instance, methods like Zero-1-to-3 \cite{Liu2023zero1to3,shi2023zero123++}, EscherNet \cite{kong2024eschernet}, CAT3D \cite{gao2024cat3d} and SEVA \cite{zhou2025stable} fine-tune 2D image generative models \cite{rombach2022high} on multi-view datasets to apply them to the NVS task. However, since these 2D models inherently lack cross-view perception, their priors provide limited help in inferring occluded regions and do not contribute to 3D consistency.
While CAT4D \cite{wu2025cat4d} and AC3D \cite{bahmani2025ac3d} utilize video models and incorporate camera conditioning for NVS, they are primarily designed for general video generation, where objects may move or transform over time. Consequently, they achieve suboptimal performance on the object-centric NVS task we focus on.
In contrast, SV3D \cite{voleti2024sv3d} pioneers a different paradigm by fine-tuning a pre-trained video generation model \cite{blattmann2023stable} to leverage its rich visual commonsense for novel view synthesis of static objects. Nevertheless, due to its relatively simple camera conditioning mechanism and training strategy, the method exhibits limitations in camera control accuracy, geometric consistency, and appearance clarity, ultimately failing to fully realize the potential of this paradigm.

To address these issues, we propose OrbitNVS, a framework that reframes NVS as an orbital video generation task. Our approach incorporates a leading video generation model and introduces tailored camera conditioning and training strategies to unleash its rich visual commonsense for NVS.
Specifically, 
drawing inspiration from post-training strategies for video models \cite{he2024cameractrl,bahmani2025ac3d}, we integrate multiple camera adapters into the model to condition the generation process on the camera trajectory. These adapters modulate the feature maps in a pixel-wise manner based on the input camera pose, enabling effective camera control. 
Since normal maps have demonstrated their effectiveness in helping models understand 3D geometric structures \cite{long2024wonder3d,li2024era3d}, we introduce an auxiliary normal map generation branch and facilitate feature interaction between the two branches via self-attention mechanisms. This allows the normal map to guide the RGB synthesis, resulting in more coherent geometric structures. 
Furthermore, to improve fine appearance details, we incorporate a pixel-space loss that maps latent representations back to the pixel domain for direct supervision, thereby mitigating detail loss or blurring caused by resolution compression in latent spaces.

In summary, our contributions are as follows:
\begin{itemize}
\item[1)] We propose OrbitNVS, a novel framework that incorporates a pre-trained video generation model into NVS via carefully designed camera conditioning and training mechanisms, effectively leveraging its rich visual commonsense for high-quality view synthesis.
\item[2)] To enhance both geometric and appearance fidelity, we introduce two dedicated techniques: a normal map generation branch to reinforce 3D coherence, and a pixel-space training loss to recover fine-grained texture details.
\item[3)] Extensive experiments demonstrate that OrbitNVS achieves state-of-the-art performance in NVS, particularly in the challenging single-image setting. For instance, under a level camera orbit, it improves PSNR by +2.9 dB and +2.4 dB on the GSO and Omniobject3D benchmarks, respectively.
\end{itemize}

\section{Related Work}
\label{sec:related_work}

\textbf{Novel View Synthesis.}
Existing NVS methods can be categorized into three main paradigms based on their underlying 3D representations.
The first category relies on \textbf{Explicit 3D Representations}. Methods like \cite{charatan2024pixelsplat,held2025triangle} directly optimize representations such as NeRF \cite{mildenhall2020nerf} or 3DGS \cite{kerbl20233dgs} via differentiable rendering from dense views. While effective with dense inputs, they struggle in sparse-view settings. Generalization efforts involve depth cues \cite{yu2021pixelnerf,deng2022depthnerf,szymanowicz2024splatterimg}, regularization \cite{niemeyer2022regnerf,yang2023freenerf,jain2021nerfonadiet}, or SDS loss \cite{sargent2023zeronvs,poole2022dreamfusion,roessle2024l3dg}. However, their fundamental limitation remains: optimizing a separate 3D representation per scene impedes the transfer of multi-view knowledge across objects.
The second category involves \textbf{Intermediate 3D Caches}. Methods such as \cite{huang2025voyager,yu2024viewcrafter,ren2025gen3c} first reconstruct a 3D cache \cite{wang2024dust3r,wang2025vggt} (e.g., a coarse point cloud) from reference inputs, then render it from the target viewpoint to condition a diffusion model. While this incorporates 3D consistency, the two-stage process is computationally costly and prone to error accumulation, where reconstruction inaccuracies degrade the final output. Moreover, such caches lack the inherent visual common sense needed for reasoning about unobserved areas.
The third category, \textbf{Implicit 3D Priors}, learns 3D common sense from multi-view image data, embedding 3D knowledge into generative model parameters. Early methods achieved single novel view synthesis \cite{jin2024lvsm, liu2023zero123}, while recent advances enable multi-view generation \cite{kong2024eschernet, liu2023syncdreamer, ye2023consistent123, zheng2024free3d}. These approaches typically fine-tune 2D generative models, but their lack of cross-frame reasoning limits 3D consistency. Newer works like ViVid-1-to-3 \cite{kwak2024vivid123} and SV3D \cite{voleti2024sv3d} use video diffusion models for stronger spatial reasoning. Nevertheless, with relatively simple architectures and training strategies, these methods still underperform and have yet to fully realize this paradigm's potential.

\textbf{Video Generation.}
Benefiting from the development of diffusion generation paradigms \cite{ho2020denoising,rombach2022high,dai2024one,dai2025beyond} and transformer model architectures \cite{vaswani2017attention,liang2025order,dai2023disentangling}, video generation models have achieved impressive fidelity \cite{peebles2023dit,blattmann2023svd}. However, their application to NVS presents challenges.
Firstly, although these models offer diverse conditioning options—e.g., text \cite{blattmann2023svd, wang2024magicvideo, chen2024videocrafter2, zheng2024opensora, zhou2025scaling}, image \cite{wan2025wan, yangcogvideox, jiang2024videobooth, jiang2025vace, liu2025phantom, fei2025skyreelsa2}, speech \cite{wan2025wan}, or customization \cite{zhou2024sugar,zhang2015kaleido,wang2024customvideo}—NVS requires more flexible and precise camera viewpoint control. 
Specifically, some existing methods \cite{wan2025wan, kwak2024vivid123} only support relatively simple camera motion patterns, such as panning or zooming, which are inadequate for the arbitrary novel views or complex trajectories required in NVS. Although works such as CAT4D \cite{wu2025cat4d} and AC3D \cite{bahmani2025ac3d} introduce dedicated camera adapters or specialized architectures to achieve more flexible camera control, they are primarily tailored for general video generation. In these scenarios, objects often undergo temporal motion or transformations over time, leading to suboptimal performance on the static object NVS task that we focus on.
Secondly, conventional video generation prioritizes temporal continuity and visual plausibility \cite{zheng2024opensora,wan2025wan,wang2024magicvideo,zhao2025hunyuan3d}, often overlooking inherent 3D geometric consistency, paramount for NVS \cite{voleti2024sv3d}. This is evident in orbital videos, where models struggle to maintain consistent object geometry and scene structure across changing viewpoints. Therefore, substantial adaptation is required to meet NVS's stringent demands for flexible viewpoint control and 3D consistency.

\section{Method}
\label{sec:method}

In this section, we first describe our problem and define the notation (\S \ref{subsec:method-problem}), then introduce the video generation model we employ (\S \ref{subsec:method-foundation}), and finally, detail the model design and training strategy of our proposed OrbitNVS (\S \ref{subsec:method-design}).

\subsection{Problem Formulation}
\label{subsec:method-problem}

We formulate the NVS task as an orbital video generation problem: given a camera trajectory and a few known reference frames on it, the goal is to generate the complete orbital video. This can be expressed as:
\begin{align}
\mathcal{V}^{1:T} \sim \mathcal{P}\left(\mathcal{V}^{1:T}|\mathcal{F}^{R},\mathcal{C}^{1:T}\right),
\end{align}
where $\mathcal{V}^{1:T}$ denotes the complete orbital video, $\mathcal{F}^{R}$ is the reference frame images, $\mathcal{C}^{1:T}$ is the camera trajectory, and $T$ is the total number of frames, which we set to 61. While our formulation primarily assumes a continuous camera trajectory as input, the proposed method remains compatible with discrete viewpoints by interpolating them into a continuous trajectory.

From the perspective of controllable video generation, OrbitNVS is analogous to a video inpainting model. Specifically, it completes a masked video where only a few reference frames are provided, conditioned on the camera trajectory. In our setting, the first frame is always provided as a reference frame.

\subsection{Foundation Model}
\label{subsec:method-foundation}

To leverage a pre-trained model's prior knowledge for extrapolating unseen views and its temporal consistency for spatial coherence, we adopt the leading video generation model, Wan2.1-I2V-14B \cite{wan2025wan}, as our foundation model.

\textbf{Framework.} 
The model follows the latent diffusion paradigm \cite{rombach2022high,liang2024diffusion}, first encoding the video into a latent representation, $x_{1}=\mathcal{E}(\mathcal{V}^{1:T})$, where $\mathcal{E}$ is the VAE \cite{kingma2013auto} encoder. 
Subsequently, it applies a diffusion and denoising process within this space. The denoising is optimized using flow matching \cite{lipman2023flow,esser2024scaling}, which involves constructing a path from noise to the latent representation and training the model to predict the velocity at any point along this trajectory. Specifically, we employ rectified flow \cite{liu2023flow}, where the path is defined as a straight line. The expression is as follows:
\begin{align}
x_{t} &= tx_{1} + (1-t)x_{0}, \\
v_{t} &= \frac{dx_{t}}{dt} = x_{1} - x_{0}.
\end{align}
Here, $x_{t}$ represents the noised latent, $x_0 \in \mathcal{N}(0,1)$ is a random noise, $t \in [0,1]$ represents the timestep, and $v_t$ represents the ground truth velocity. The loss function can be formulated as the mean squared error between the model output and $v_t$ at any point along the trajectory:
\begin{align}
\mathcal{L}_{latent} :=  ||v_t-\hat{v}_{\theta}(x_t,t)||_{2}^{2}.
\end{align}
Here, $\hat{v}_{\theta}$ denotes the model's velocity prediction function.

\begin{figure*}[t]
    \centering
    \includegraphics[width=1.00\textwidth]{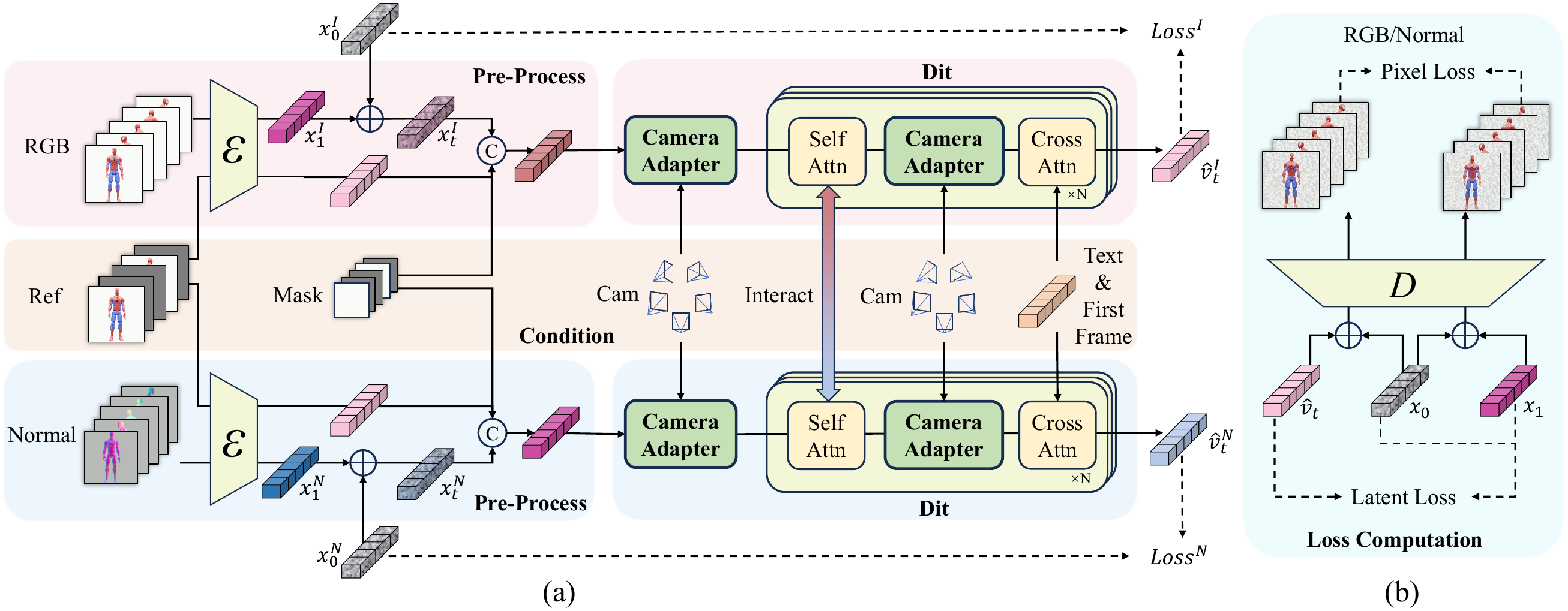}
    \caption{Overview of OrbitNVS.
(a) Model architecture. The model incorporates camera conditions by inserting camera adapters before the first and into each subsequent Dit block. A normal map generation branch, which shares parameters with the primary branch, is added to guide the RGB video synthesis. Other conditions are injected via two pathways: 1) channel-wise concatenation of the reference frame's latent representation and a positional mask with the noised latent, and 2) integration of text embeddings and the first frame's CLIP embedding via cross-attention at each layer.
(b) Loss computation. The latent representations are decoded back to pixel space using the VAE decoder to compute a pixel-space loss, while the original latent loss is retained.}
    \label{fig:overview}
\end{figure*}

\textbf{Conditions.} 
This model generates a video from an initial frame and a text prompt. To inject the first frame as a condition, it follows the approach from works \cite{zhang2023i2vgen,blattmann2023stable,yangcogvideox}: the frame is concatenated with zero-filled frames and fed into the VAE encoder to obtain a conditional latent. This conditional latent, along with a binary mask indicating which frames are conditioned, is then concatenated with the original noised latent along the channel dimension to guide the denoising process. Additionally, within each layer of the Diffusion Transformer (DiT) \cite{peebles2023scalable}, the model injects text embeddings and the CLIP \cite{radford2021learning} image embedding of the first frame via cross-attention.

\subsection{Model Design and Training}
\label{subsec:method-design}

Although the foundation model possesses extensive world knowledge, it faces the following challenges when applied to the task of orbital video generation: 1) It cannot interpret camera pose inputs. 2) As a general-purpose model, it struggles with generating plausible and consistent 3D shapes and textures. To address these challenges, we first insert a camera adapter into the base model, endowing it with the ability to perceive camera information. Then, to adapt the model for the orbital video generation task, we propose two training strategies targeting the key attributes of shape plausibility and appearance clarity: normal map generation and pixel-space post-training. We will elaborate on these in the following sections.

\textbf{Camera Adapter.}
We use Plücker coordinates to represent camera poses. Specifically, for an orbital video $\mathcal{V}^{1:T} \in \mathbb{R}^{3 \times T \times H \times W}$ with $T$ frames, height $H$, width $W$, and $3$ color channels, we use the given camera poses $\mathcal{C}^{1:T}$ to compute the Plücker coordinates (a 6D vector) for the imaging ray of each pixel. This process yields a Plücker tensor $\hat{\mathcal{C}}^{1:T} \in \mathbb{R}^{6 \times T \times H \times W}$
, which shares the same spatial and temporal dimensions as the video but differs in the channel dimension.
Following works \cite{he2024cameractrl,wan2025wan}, we insert camera adapters preceding the first layer of the DiT model and between the self-attention and cross-attention blocks within each layer, as shown in Figure \ref{fig:overview} (a). 
These adapters take the Plücker tensor $\hat{\mathcal{C}}^{1:T}$ as input to predict the scale and shift parameters for Adaptive Layer Normalization (AdaLN) \cite{peebles2023scalable}. 
The camera condition is thereby incorporated by modulating the distribution of the original DiT features via the AdaLN mechanism, as formulated below:
\begin{align}
\mu, \sigma = \mathcal{A}(\hat{\mathcal{C}}^{1:T}),\\
\hat{z} = (1+\sigma) \cdot z + \mu.
\end{align}
Here, $\hat{z}$ represents the features infused with camera conditions, $z$ denotes the original DiT features, $\mathcal{A}$ refers to the camera adapter, and $\mu$ and $\sigma$ represent the predicted shift and scale parameters, respectively. 
The camera adapter is composed of two convolutional layers with parallel residual connections, followed by an output convolutional layer that is zero-initialized to preserve the original DiT feature distribution at the start of training.

\textbf{Normal Map Generation.}
The optimization objective of existing diffusion models is to directly align the appearance of the generated video with the ground-truth video, which is jointly determined by geometry and texture. This entanglement implies that, from the model’s perspective, two objects sharing the same geometric but differing in texture are considered fundamentally distinct. As a result, the model struggles to directly learn an object’s geometric features.
To address this issue, we introduce a normal map generation branch that shares parameters with the main branch, as illustrated in Figure \ref{fig:overview} (a). This branch interacts with the main branch at the self-attention layers, guided by the same Plücker embeddings. Through this design, OrbitNVS is able to simultaneously generate pixel-aligned RGB maps and normal maps. This approach offers dual benefits: on one hand, it allows geometric cues from the normal maps to inform the RGB generation process, leading to improved geometric consistency. On the other hand, by incorporating supervision from the normal maps, the model can directly perceive object shape independent of texture variations, thereby facilitating more effective learning of geometric structures.

\textbf{Pixel-Space Post-Training.}
The existing latent diffusion paradigm supervises the denoising model in the latent space. While this significantly reduces computational costs and accelerates training, it comes at the cost of lossy compression, which can lead to the inaccuracy or loss of high-frequency details \cite{zhang2024pixel}.
To address this issue, we introduce a Pixel-Space Post-Training stage. In this stage, we decode the latent variables back into pixel space using the VAE's decoder and then perform the alignment, as shown in Figure \ref{fig:overview} (b). Despite the inherent information loss during VAE compression, our direct supervision in pixel space allows the DiT to be trained via gradients back-propagated through the VAE decoder. This process enables the DiT to learn to produce latent representations that are better optimized for the decoder, ultimately resulting in high-fidelity video generation. The objective is expressed as follows:
\begin{align}
\mathcal{L}_{pixel} := ||\mathcal{D}(tx_{1} + (1-t)x_0)-\mathcal{D}(x_0 + t\hat{v}_{\theta}(x_{t},t))||_{2}^{2}.
\end{align}
Here, $x_{1} \in \mathcal{E}(\mathcal{V})$ represents the latent obtained by encoding the video with the VAE encoder, $x_0 \in \mathcal{N}(0,1)$ is noise sampled from a normal distribution, and $t \in [0,1]$ denotes the timestep. $x_{t} = tx_{1} + (1-t)x_0$ is the noised latent, and $\hat{v}_{\theta}(x_{t},t)$ represents the velocity predicted by the model.

Considering the high computational complexity of this loss function, we adopt a two-stage training strategy. In the first stage, we train the model using only the standard latent space supervision. Then, in the second stage, we take the model from the first stage and continue training it with joint supervision from both the latent and pixel spaces.

\section{Experiments}
\label{sec:experiments}

\subsection{Settings}
In this section, we present our experimental setup in three aspects: the benchmarks and metrics (\S \ref{subsubsec:benchmarks}), baseline methods for comparison (\S \ref{subsubsec:baseline}), and implementation details of OrbitNVS (\S \ref{subsubsec:implementation}).

\subsubsection{Benchmarks and Metrics.}\label{subsubsec:benchmarks}
Following SV3D \cite{voleti2024sv3d}, we use Omniobject3D \cite{wu2023omniobject3d} and GSO \cite{downs2022google}, unseen during model training, as our test datasets. Omniobject3D comprises 6k 3D assets, and GSO contains 1k 3D assets, both acquired by scanning real-world objects. To compare the NVS performance of different methods, we render the 3D assets into orbital videos. The initial rendering resolution is 1024 $\times$ 1024 with 61 frames, which are subsequently resized to the required resolution for different methods. The camera trajectory for our rendering involves the azimuth uniformly circling the object, while the elevation angle fluctuates sinusoidally, as expressed below:
\begin{align}
a_i &= \frac{2\pi \cdot i }{T_{val}}, \label{eq:azimuth} \\
e_i &= \mathcal{U}_{val} \cdot sin(f\cdot\frac{2\pi \cdot i }{T_{val}}). \label{eq:elevation}
\end{align}
Here, $a_i$ and $e_i$ represent the azimuth and elevation of the camera for the $i$-th frame, respectively. $T_{val}$ denotes the number of frames, set to 61. $f$ represents the frequency of the elevation's fluctuation, set to 3, and $\mathcal{U}_{val}$ indicates the amplitude of this fluctuation. We established three levels of difficulty by setting three different amplitudes for $\mathcal{U}_{val}$: 0$^{\circ}$, 30$^{\circ}$, and 60$^{\circ}$. 
An amplitude of 0$^{\circ}$ signifies a level, circular camera orbit.
Additionally, we ensure the camera always points towards the object's center, normalize the object's size, and set the camera's rotation radius to 4 to maintain a moderate object size in the video.

To evaluate the similarity between the generated videos and the real videos, we use image similarity metrics: Peak Signal-to-Noise Ratio (PSNR), Structural SIMilarity (SSIM), and Learned Perceptual Similarity (LPIPS) \cite{zhang2018unreasonable}.

\subsubsection{Baselines}\label{subsubsec:baseline}
We use three current SOTA works in NVS as our baselines: SV3D \cite{voleti2024sv3d}, SEVA \cite{zhou2025stable}, and EscherNet \cite{kong2024eschernet}. Among them, SV3D's paradigm is the most similar to ours, as it was the first to frame the NVS task as an orbital video generation task.
To ensure a proper reproduction, we adapt the video resolution and number of frames to meet the input requirements of each baseline. For instance, SV3D requires 21 frames, so we uniformly downsample our 61-frame rendered videos. The other two baselines utilize the full 61 frames. Additionally, the videos are resized to the resolutions required by each baseline: 576$\times$576 for SV3D and SEVA, and 256$\times$256 for EscherNet.

\begin{figure*}[t]
    \centering
    \includegraphics[width=1.00\textwidth]{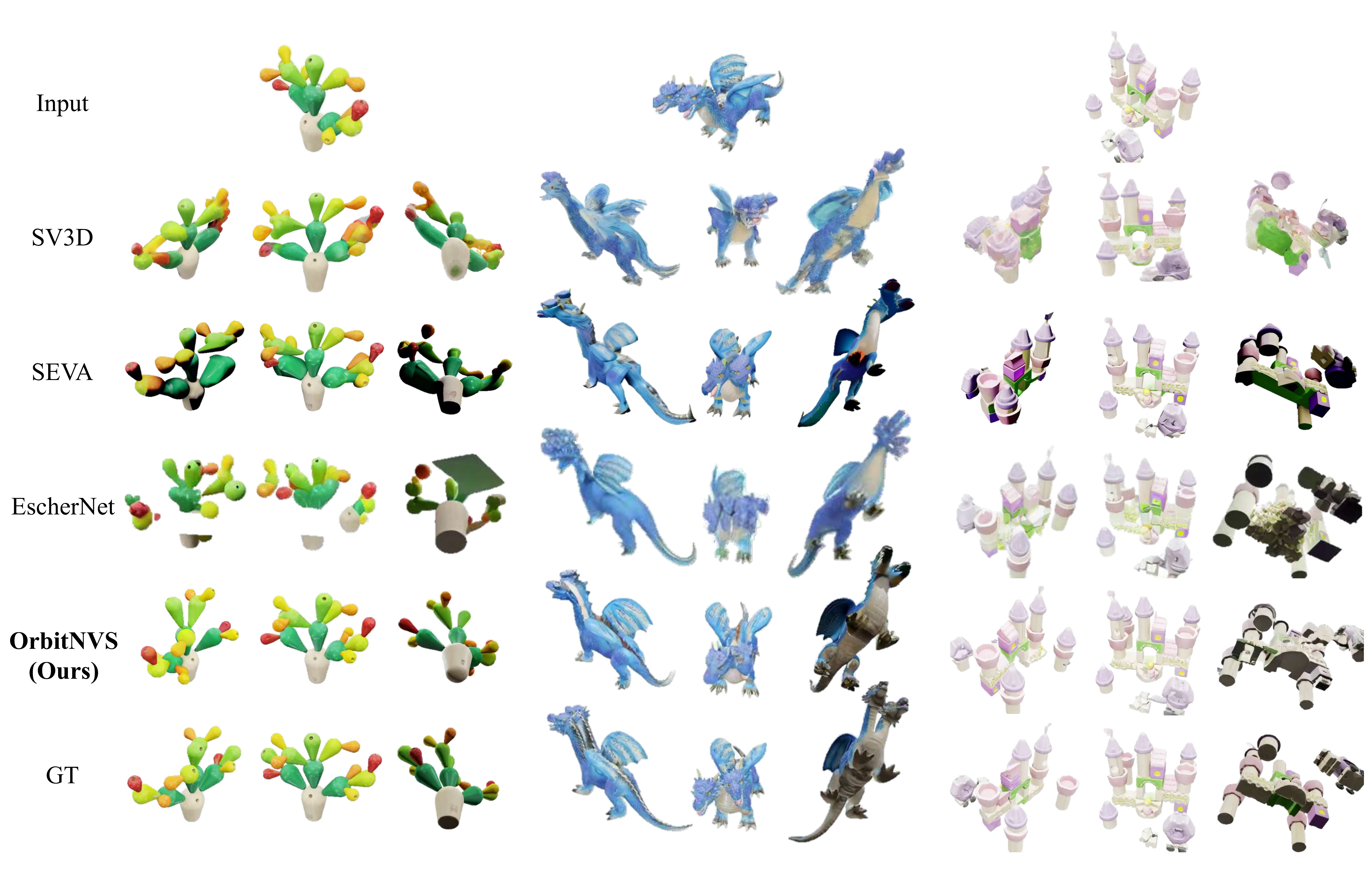}
    \caption{Qualitative comparison of NVS results generated by different methods under the single reference view setting.}
    \label{fig:single_view}
\end{figure*}

\begin{table*}[h]
\centering
\caption{Experimental results on the GSO and OmniObject3D benchmarks under a single reference view and different camera trajectories. The values 0$^{\circ}$, 30$^{\circ}$, and 60$^{\circ}$ denote the elevation oscillation amplitude of the sinusoidal camera trajectory (0$^{\circ}$ represents a level orbit). ``w/o normal'' and ``w/o pixel-loss'' are ablations of our OrbitNVS without the pixel-space loss and normal map branch, respectively. Each entry reports the ``PSNR$\uparrow$ SSIM$\uparrow$ LPIPS$\downarrow$'' metrics, with the best results highlighted in \textbf{bold}.}
\newcommand{\myTableWidth}{11cm} 
\newcommand{\firstColWidth}{2.7cm}


\begin{tabular*}{\myTableWidth}{@{\extracolsep{\fill}}p{\firstColWidth}|ccc|ccc|ccc}
\toprule
\multicolumn{1}{l}{\textbf{GSO}} & \multicolumn{3}{c}{$0^{\circ}$} & \multicolumn{3}{c}{$30^{\circ}$} & \multicolumn{3}{c}{$60^{\circ}$} \\
\midrule
SEVA & 18.3 & 0.87 & 0.15 & 16.3 & 0.84 & 0.17 & 13.7 & 0.80 & 0.24 \\
EscherNet & 20.8 & 0.85 & 0.15 & 19.8 & 0.84 & 0.15 & 19.3 & 0.83 & 0.17 \\
SV3D & 18.9 & 0.88 & 0.18 & 17.5 & 0.86 & 0.22 & 17.6 & \textbf{0.86} & 0.22 \\
\midrule
\textbf{OrbitNVS} & \textbf{23.7} & \textbf{0.89} & \textbf{0.11} & \textbf{21.9} & \textbf{0.87} & \textbf{0.13} & \textbf{21.0} & \textbf{0.86} & \textbf{0.15} \\
w/o normal & 22.3 & \textbf{0.89} & 0.12 & 20.8 & \textbf{0.87} & 0.14 & 19.9 & \textbf{0.86} & 0.16 \\
w/o pixel-loss & 22.8 & \textbf{0.89} & \textbf{0.11} & 21.2 & \textbf{0.87} & \textbf{0.13} & 20.0 & 0.85 & 0.16 \\
\bottomrule
\end{tabular*}

\vspace{1.5em} 

\begin{tabular*}{\myTableWidth}{@{\extracolsep{\fill}}p{\firstColWidth}|ccc|ccc|ccc}
\toprule
\multicolumn{1}{l}{\textbf{OmniObject3D}} & \multicolumn{3}{c}{$0^{\circ}$} & \multicolumn{3}{c}{$30^{\circ}$} & \multicolumn{3}{c}{$60^{\circ}$} \\
\midrule
SEVA & 16.7 & 0.83 & 0.19 & 16.1 & 0.82 & 0.19 & 15.9 & 0.82 & 0.19 \\
EscherNet & 18.6 & 0.82 & 0.19 & 18.2 & 0.81 & 0.19 & 18.1 & 0.81 & 0.20 \\
SV3D & 16.4 & 0.84 & 0.24 & 16.7 & 0.84 & 0.23 & 17.4 & \textbf{0.85} & 0.21 \\
\midrule
\textbf{OrbitNVS} & \textbf{21.0} & \textbf{0.87} & \textbf{0.14} & \textbf{20.4} & \textbf{0.86} & \textbf{0.14} & \textbf{19.9} & \textbf{0.85} & \textbf{0.16} \\
w/o normal & 20.1 & 0.85 & 0.16 & 19.1 & 0.85 & 0.17 & 18.6 & 0.84 & 0.18 \\
w/o pixel-loss & 20.8 & 0.86 & \textbf{0.14} & 19.7 & 0.84 & 0.15 & 19.7 & 0.84 & \textbf{0.16} \\
\bottomrule
\end{tabular*}
\label{tab:single_view}
\end{table*}

\subsubsection{Implementation Details}\label{subsubsec:implementation}
Our training data consists of 320k assets from the Step1X-3D dataset \cite{li2025step1x}, which were filtered from the 800k assets in the Objaverse-V1 \cite{deitke2023objaverse}.
For each asset, we rendered a 129-frame orbital video and its corresponding normal maps at a 448$\times$448 resolution. 
To ensure viewpoint diversity, we randomly employed one of two camera motion modes during rendering: the first was a random mode, where the camera orientation was adjusted with random offsets at each step, and the second was a sinusoidal oscillation mode, similar to equations \ref{eq:azimuth} and \ref{eq:elevation}, but with random biases, frequencies, amplitudes, and initial phases.
From each generated video, we randomly sampled 61 frames for training to further enhance the diversity of camera motion. 
For text input, during training, we used Qwen2.5-VL-7B \cite{bai2025qwen2} to generate detailed captions describing the object in the video. At test time, captions were generated for the reference images to form the text prompt.
Our training process was divided into two stages. In the first stage, we trained the model for 3200 steps using only a latent space loss, which required approximately 60 hours. In the second stage, we fine-tuned the model for an additional 1800 steps using a combination of latent and pixel space losses, which took an additional 66 hours. 
The model was trained with a total batch size of 64 on a cluster of 8 machines, each equipped with 8 NVIDIA A800 (80G VRAM) GPUs. We adopted a grouped learning rate strategy, applying a learning rate of $1e^{-5}$ to parameters loaded from pre-trained weights and $1e^{-4}$ to newly initialized parameters, such as the camera adapter.


\begin{table*}[h]
\centering
\caption{Experimental results of different NVS methods under a level camera trajectory (0$^{\circ}$ elevation) with varying numbers of reference views. Each entry reports the ``PSNR$\uparrow$ SSIM$\uparrow$ LPIPS$\downarrow$'' metrics, with the best results highlighted in \textbf{bold}.}
\newcommand{\myTableWidth}{10.8cm} 
\newcommand{\firstColWidth}{2.7cm}


\begin{tabular*}{\myTableWidth}{@{\extracolsep{\fill}}p{\firstColWidth}|ccc|ccc|ccc}
\toprule
\multicolumn{1}{l}{\textbf{GSO}} & \multicolumn{3}{c}{1 ref.} & \multicolumn{3}{c}{2 ref.} & \multicolumn{3}{c}{3 ref.} \\
\midrule
SV3D & 18.9 & 0.88 & 0.18 &  & - &  &  & - &  \\
SEVA & 18.3 & 0.87 & 0.15 & 25.7 & \textbf{0.92} & 0.07 & 28.3 & \textbf{0.94} & \textbf{0.04} \\
EscherNet & 20.9 & 0.85 & 0.15 & 23.4 & 0.87 & 0.13 & 26.3 & 0.91 & 0.06 \\
\midrule
\textbf{OrbitNVS} & \textbf{23.7} & \textbf{0.89} & \textbf{0.11} & \textbf{26.5} & \textbf{0.92} & \textbf{0.06} & \textbf{28.8} & \textbf{0.94} & \textbf{0.04} \\
\bottomrule
\end{tabular*}

\vspace{1.5em} 

\begin{tabular*}{\myTableWidth}{@{\extracolsep{\fill}}p{\firstColWidth}|ccc|ccc|ccc}
\toprule
\multicolumn{1}{l}{\textbf{OmniObject3D}} & \multicolumn{3}{c}{1 ref.} & \multicolumn{3}{c}{2 ref.} & \multicolumn{3}{c}{3 ref.} \\
\midrule
SV3D & 16.4 & 0.84 & 0.24 &  & - &  &  & - &  \\
SEVA & 16.7 & 0.83 & 0.19 & 23.9 & \textbf{0.90} & 0.09 & 27.1 & \textbf{0.92} & \textbf{0.05} \\
EscherNet & 18.6 & 0.82 & 0.19 & 22.8 & 0.86 & 0.10 & 25.3 & 0.89 & 0.06 \\
\midrule
\textbf{OrbitNVS} & \textbf{21.0} & \textbf{0.87} & \textbf{0.14} & \textbf{24.4} & \textbf{0.90} & \textbf{0.08} & \textbf{27.3} & \textbf{0.92} & \textbf{0.05} \\
\bottomrule
\end{tabular*}
\label{tab:multi_view}
\end{table*}

\subsection{Main Results}
We evaluate our method against baselines from three perspectives:
1) Reasoning on unseen views: To validate the model's capability to reason about unseen views, we provide qualitative and quantitative results under the single-view setting (Figure \ref{fig:single_view} and Table \ref{tab:single_view}).
2) Camera pose controllability: To assess the precision of camera control, we analyze performance variations across different camera trajectories (Table \ref{tab:single_view}).
3) Robustness to reference views: To examine the impact of reference views, we study how performance changes as their number increases (Table \ref{tab:multi_view}).

\textbf{Reasoning on unseen views.}
As illustrated in Figure \ref{fig:single_view}, OrbitNVS demonstrates superior performance over existing methods under the single-view setting. 
OrbitNVS's superior performance stems from two key strengths: its enhanced capability to reason about unseen regions (e.g., inferring a bottom-side view of the dragon from a top-side viewpoint) and its stronger 3D consistency (as evidenced by the toy plant and castle examples in Figure \ref{fig:single_view}).

Table \ref{tab:single_view} presents the NVS results under the single-view setting. The data demonstrate that OrbitNVS significantly outperforms all baseline methods on both the GSO and OmniObject3D benchmarks. For example, on the level camera trajectory, OrbitNVS surpasses the strongest baseline, EscherNet, by margins of +2.9 dB and +2.4 dB in PSNR on GSO and OmniObject3D, respectively. 
This indicates that OrbitNVS possesses a richer visual prior and a stronger capability to infer unseen viewpoints from a single image compared to other baselines.

\textbf{Camera pose controllability.}
Table \ref{tab:single_view} presents the NVS results under different camera orbits. The results indicate that OrbitNVS significantly outperforms all baselines across every orbit. For instance, on the GSO benchmark, when the camera elevation oscillates with amplitudes of 30$^{\circ}$ and 60$^{\circ}$, OrbitNVS achieves PSNR values that are 2.1 dB and 1.7 dB higher, respectively, than those of the best baseline, EscherNet.
This demonstrates that OrbitNVS has a more precise camera control capability than other methods.
Furthermore, we observe a monotonic decline in performance for most methods as the elevation oscillation amplitude increases. On the GSO dataset, for instance, the PSNR of OrbitNVS for the 60$^{\circ}$ trajectory is 2.7 dB lower than for the 0$^{\circ}$ trajectory, while SEVA exhibits a more substantial drop of 4.6 dB. 
In contrast, SV3D displays an anomalous, non-monotonic trend, even showing a slight improvement on OmniObject3D. We attribute this to two factors. 
First, SV3D's lower frame count (21 frames) results in sparse trajectory sampling, which may cause it to miss extreme elevation points, unintentionally simplifying the problem. 
Second, SV3D's training data includes numerous samples with large pitch variations, inherently enhancing its adaptability to such orbital patterns.

\begin{figure}[t]
    \centering
    \includegraphics[width=0.6\textwidth]{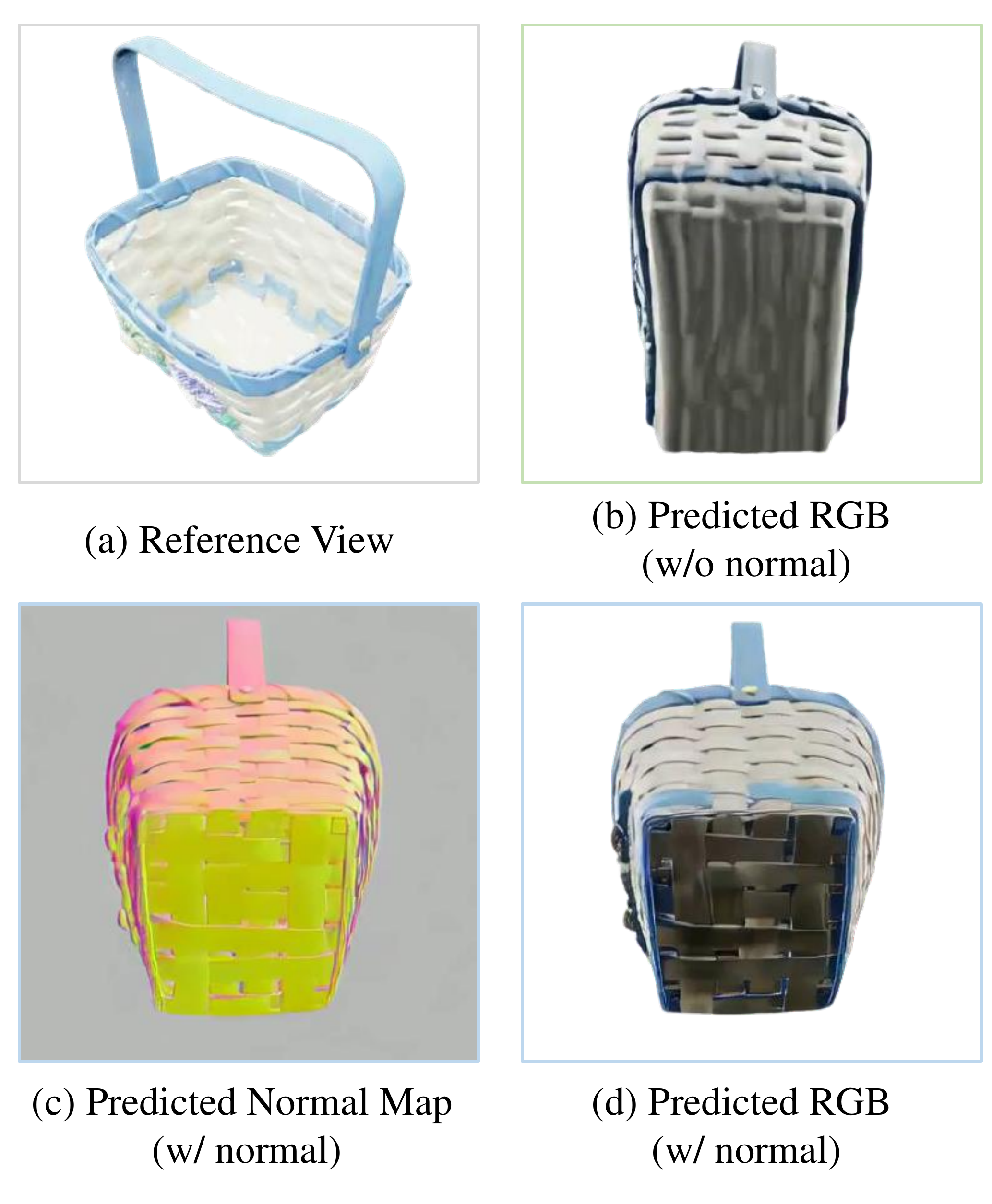}
    \caption{Ablation on the normal map generation branch. Results with the branch (c, d) show significantly clearer geometric details than those without it (b).}
    \label{fig:ablation_norm}
\end{figure}

\begin{figure}[t]
    \centering
    \includegraphics[width=0.6\textwidth]{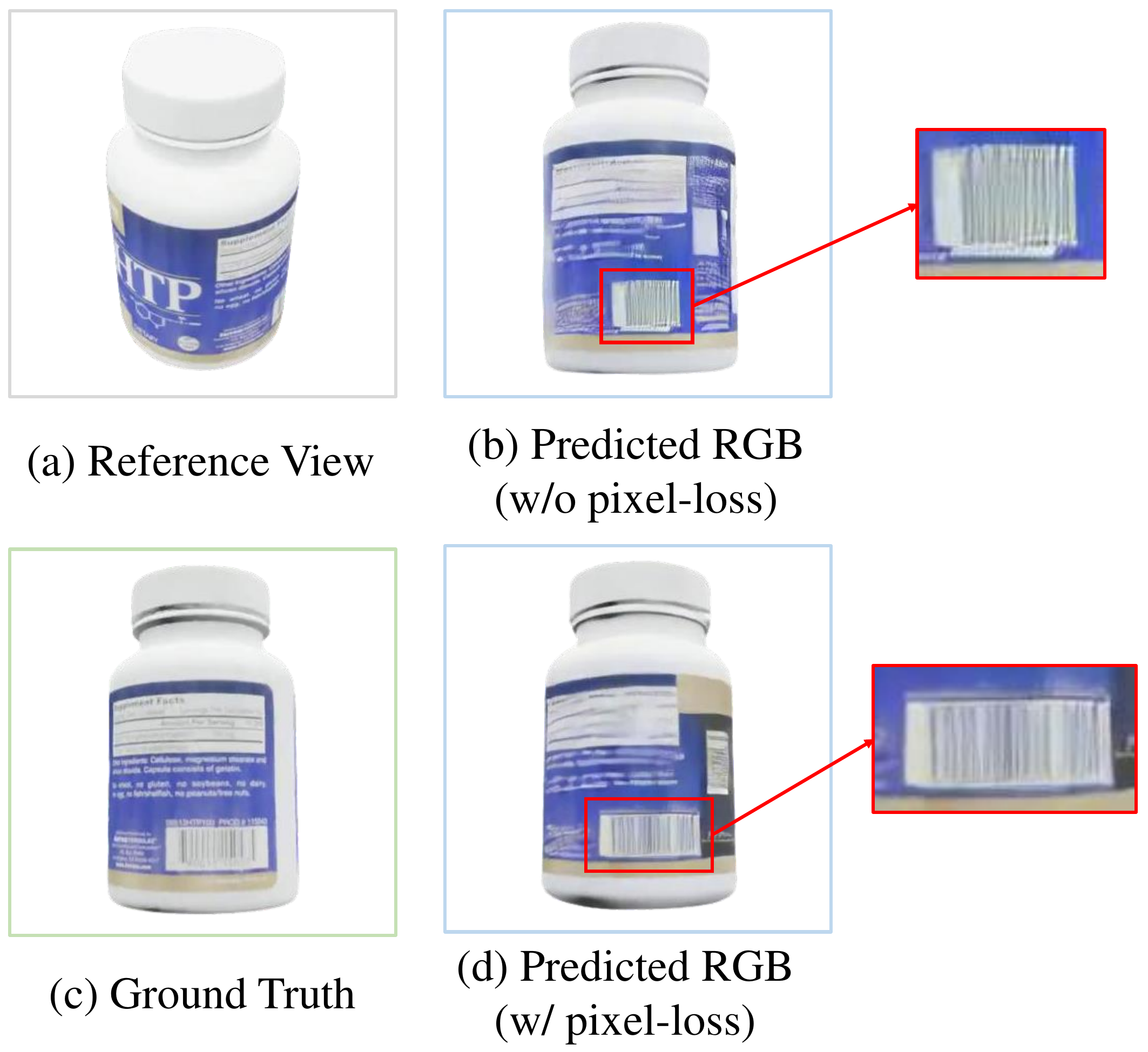}
    \caption{Ablation on the pixel-space post-training. The result with the pixel loss (d) shows superior plausibility in detailed textures compared to the result without it (b).}
    \label{fig:ablation_pixel}
\end{figure}

\textbf{Robustness to reference views.}
Table \ref{tab:multi_view} presents the impact of varying the number of reference views on NVS performance. To maximize the coverage of the visible surface, the azimuth angles of our selected reference views are uniformly distributed. For example, with three reference views, the azimuth difference between adjacent views is 120$^{\circ}$. The results in the table show that, with the exception of SV3D (which only supports single-image input), the performance of all other methods improves significantly as the number of reference views increases. When the number of views reaches three, the synthesized views already achieve a high fidelity to the ground truth. For instance, the PSNR of OrbitNVS and SEVA on GSO exceeds 28, and the SSIM reaches 0.94. This further validates our claim that the core challenge of NVS lies not in rendering minor viewpoint changes within visible areas, but in inferring the geometry and appearance of unseen views. Notably, although all methods perform well with multiple inputs, OrbitNVS still achieves the best quantitative metrics.

\subsection{Ablation}
In this section, we conduct an ablation study on our two key designs: the normal map generation branch and the pixel-space post-training.

\textbf{Impact of the normal map generation branch.}
Figure \ref{fig:ablation_norm} demonstrates the novel view synthesis (NVS) results for a bamboo basket with complex geometry. As shown in Figure \ref{fig:ablation_norm} (b), the model without the normal map generation branch fails to capture the intricate woven structure, resulting in generated novel views with missing geometric details and an overall poor shape. In contrast, with the introduction of this branch, OrbitNVS simultaneously generates both a normal map and an RGB image for the novel view that accurately reproduce the detailed geometry, as seen in Figure \ref{fig:ablation_norm} (c) and (d). The results are highly realistic, closely resembling the actual bamboo weaving. This indicates that the normal map generation branch significantly enhances the geometric fidelity of the generated orbit.
Quantitatively, the branch also leads to a substantial improvement in metrics. As presented in Table \ref{tab:single_view}, it achieves a performance gain of over 1 dB in PSNR across most settings.

\textbf{Impact of the pixel-space post-training.}
Figure \ref{fig:ablation_pixel} presents the NVS results for an object with fine-grained textures. As shown in Figure \ref{fig:ablation_pixel} (b), the model trained without the pixel-loss fails to preserve high-frequency details in areas like the barcode, resulting in blurred and irregular textures. It is important to note that this model has the same number of training steps as the model with pixel-loss, but uses only the latent loss. In contrast, the model with the pixel-loss generates a more plausible barcode, as depicted in Figure \ref{fig:ablation_pixel} (d). This demonstrates that pixel-space post-training effectively enhances the model's capability to generate high-frequency details, thereby improving visual clarity. Quantitatively, the data in Table \ref{tab:single_view} confirms that this component yields an improvement in metrics.

\subsection{Controllable NVS}
\begin{figure}[t]
    \centering
    \includegraphics[width=0.7\textwidth]{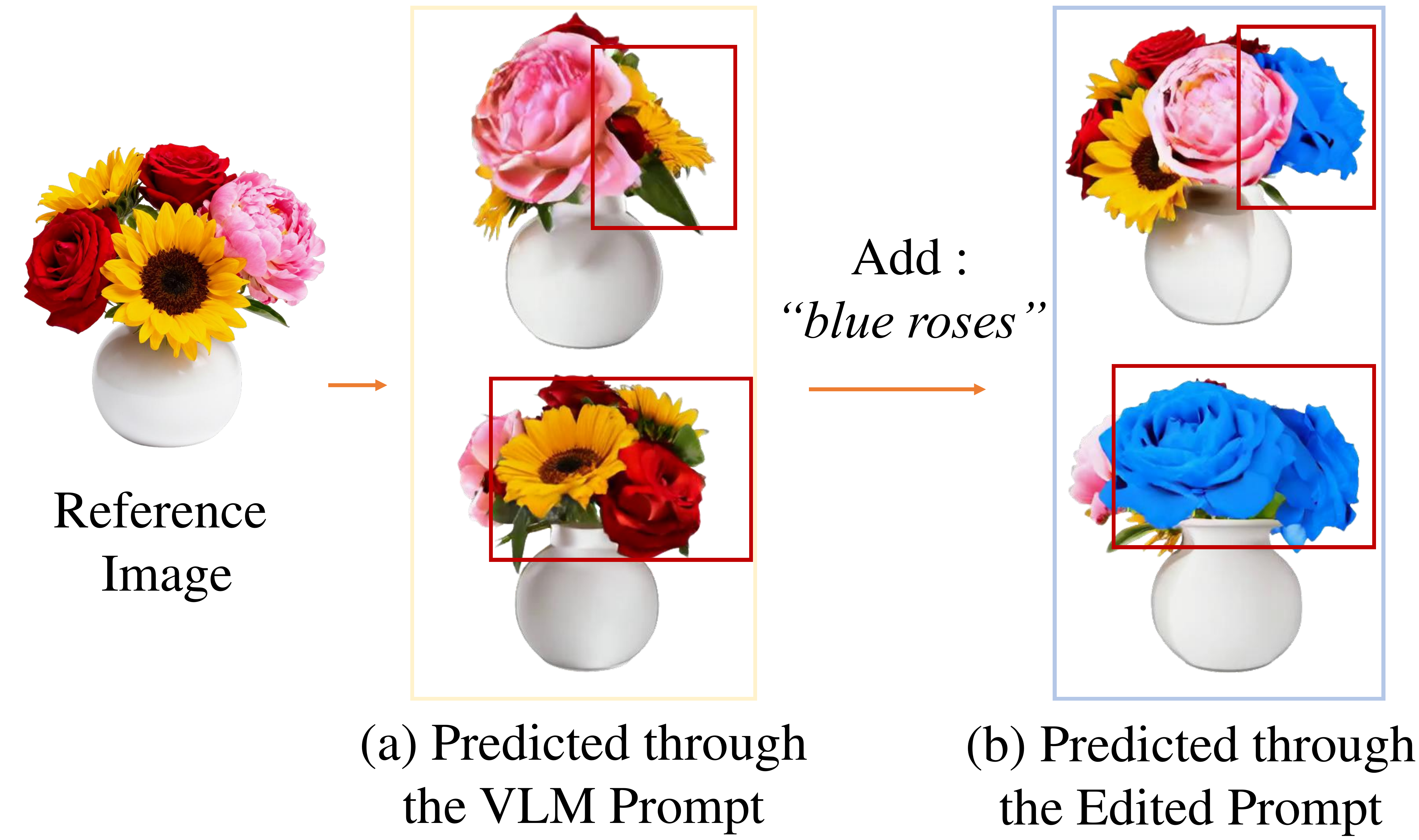}
    \caption{Controllable Novel View Synthesis is achieved by manipulating the text prompt. (a) Using a VLM to predict a prompt from the reference image as the condition for NVS. (b) Using a manually modified text prompt as the condition for NVS.}
    \label{fig:discuss_text}
\end{figure}

Conventional NVS methods rely entirely on the model's intrinsic imagination from the known view, leaving users with no control over the process. In contrast, OrbitNVS incorporates an additional text input, enabling more controllable generation through text conditions. For example, as shown in Figure \ref{fig:discuss_text} (a), if we use a caption generated by a VLM for the reference image as the text condition, the model generates a novel view with a red flower. However, if we manually edit the text prompt to include "blue roses," we can steer the model to generate a blue flower in the novel view, as shown in Figure \ref{fig:discuss_text} (b). 

\section{Conclusion}
\label{sec:conclusion}
We present OrbitNVS, which reformulates the NVS task as an orbital video generation problem, leveraging the rich visual priors of pre-trained video generation models to achieve high-quality view synthesis. Our approach introduces three main adaptations: first, by incorporating camera adapters into the video model to achieve precise camera control; second, by designing a normal map generation branch and utilizing an attention mechanism to allow normal features to guide the synthesis of novel views, thereby enhancing geometric consistency; and third, by mapping the latent representations back to the pixel space for supervision, which improves the clarity and detail of the synthesized appearance. 
Extensive experiments demonstrate that our method significantly outperforms existing methods, particularly in its superior ability to reason about unseen views and its enhanced geometric and appearance consistency.
We believe that leveraging the priors of video models to assist 3D vision tasks is a promising research avenue. OrbitNVS represents a solid step forward along this path. We will release our model to facilitate further research.

\clearpage  


%
%

\section*{Acknowledge}
This work was completed during the first author's internship at Agentic Labs, XGTech.

\bibliographystyle{splncs04}
\bibliography{main}

\clearpage

\setcounter{section}{0}
\renewcommand\thesection{\Alph{section}}
\renewcommand\thefigure{\Alph{figure}}
\renewcommand\thetable{\Alph{table}}

\begin{center}
    {\Large\textbf{Supplementary Material}} \\[4pt]
\end{center}






In the supplementary materials, we expand on three aspects: 1) We include videos for multiple examples in the supplementary archive, with an overview and analysis in \S \ref{supsec:visual}. 2) An ablation study on the text input details the role of VLM-generated captions in \S \ref{supsec:abla_text}. 3) A quantitative evaluation of the generated normal maps is presented in \S \ref{supsec:normal}.

\section{Visualization of Generated Results}\label{supsec:visual}

\begin{figure*}[h]
\centerline{\includegraphics[width=0.95\linewidth]{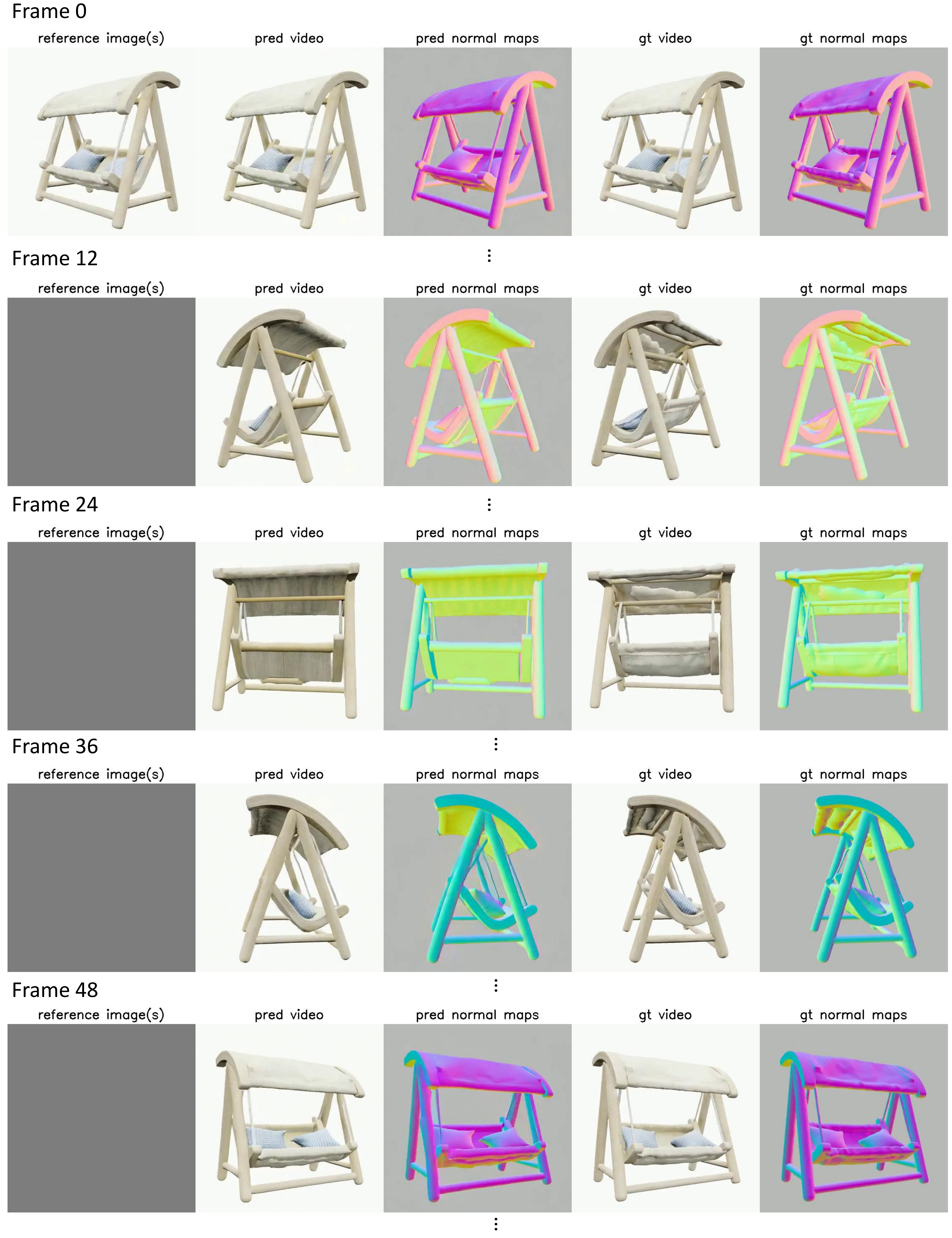}}
\caption{Example 1. A set of frames from a video generated by OrbitNVS. Each frame includes the reference image, the predicted novel-view image and normal map, and the corresponding ground-truth novel-view image and normal map.}
\label{fig:video_1}
\end{figure*}

\begin{figure*}[h]
\centerline{\includegraphics[width=0.95\linewidth]{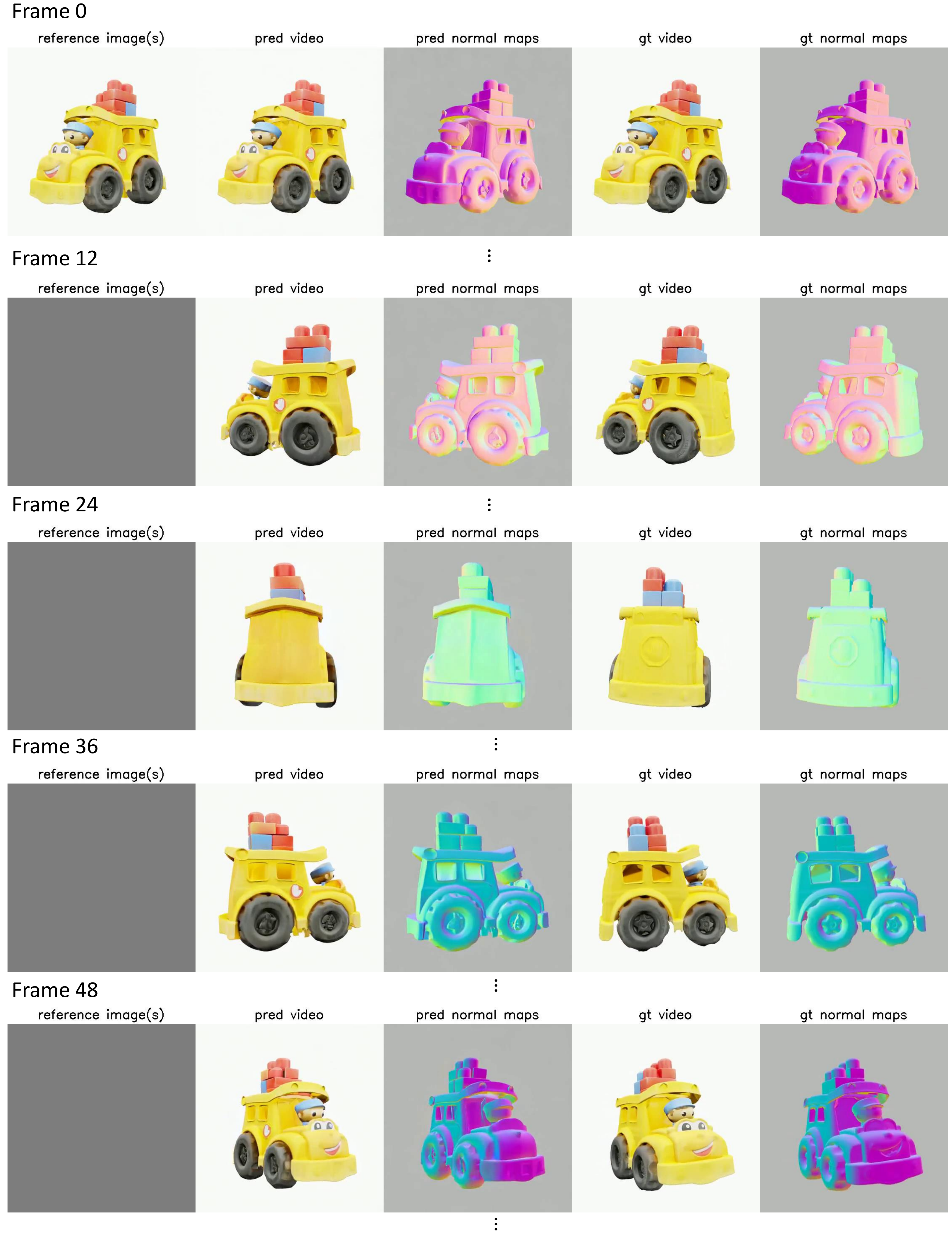}}
\caption{Example 2.}
\label{fig:video_2}
\end{figure*}

\begin{figure*}[h]
\centerline{\includegraphics[width=0.95\linewidth]{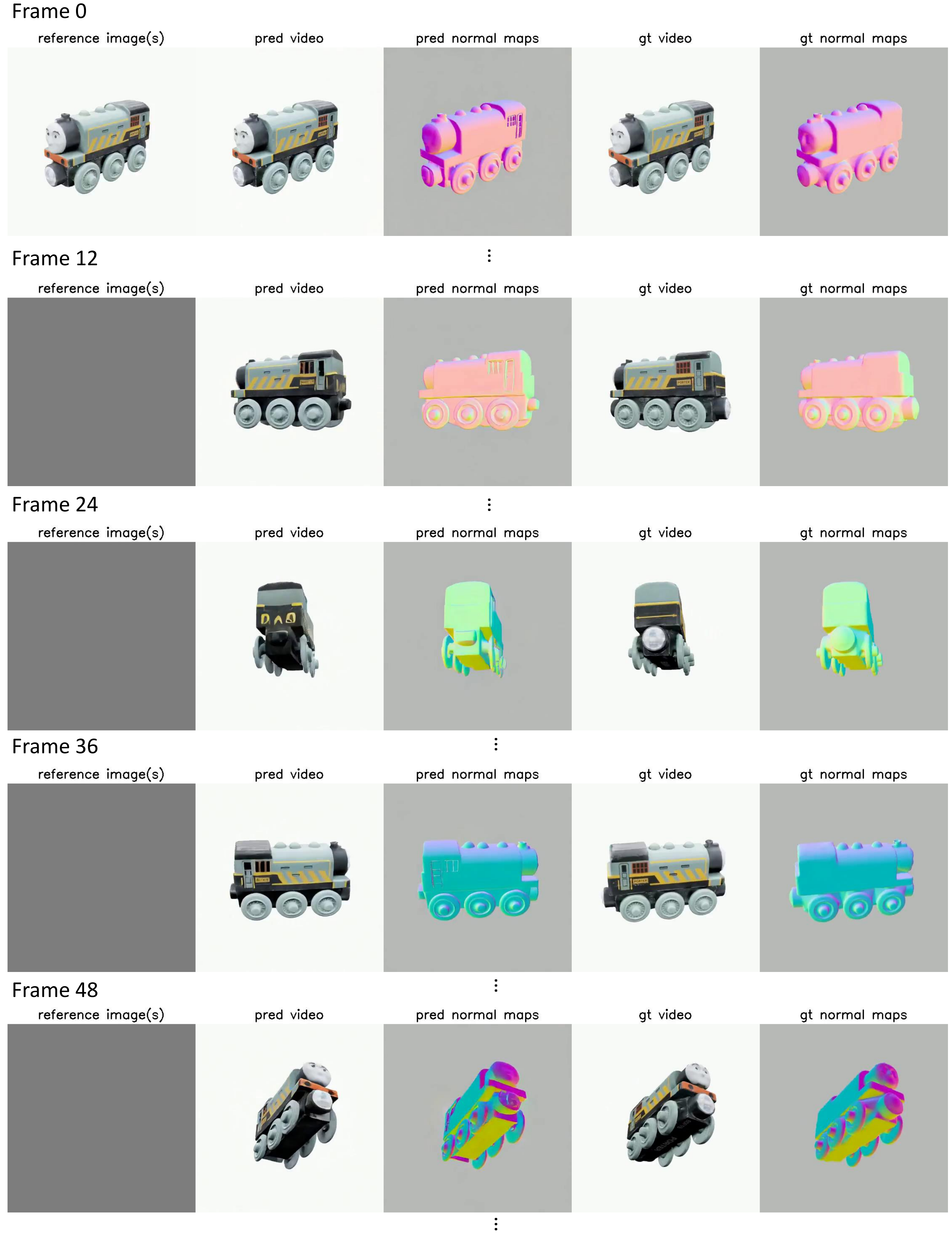}}
\caption{Example 3.}
\label{fig:video_3}
\end{figure*}

\begin{figure*}[h]
\centerline{\includegraphics[width=0.95\linewidth]{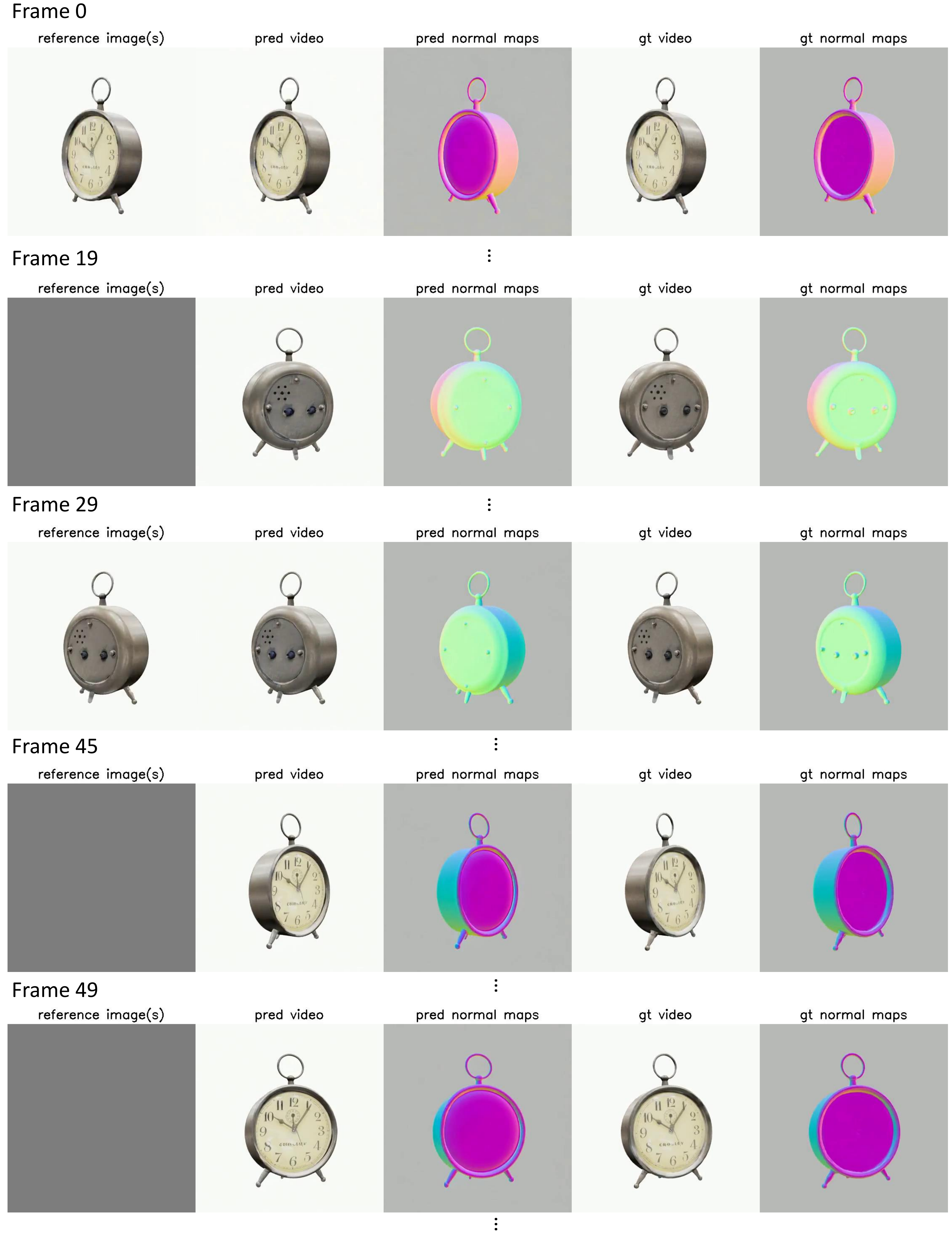}}
\caption{Example 4.}
\label{fig:video_4}
\end{figure*}

We have uploaded the complete videos of all visual results (Figures 1, 3, 4, 5, and 6) from the main text. For qualitative comparison results (Figure 3 in the main text), we have also included videos generated by baseline methods. For quantitative analysis results (Tables 1 and 2), representative generated videos are provided due to storage constraints. Several examples are shown in Figures \ref{fig:video_1}, \ref{fig:video_2}, \ref{fig:video_3}, and \ref{fig:video_4}, which showcase the input images, the generated novel-view videos and normal maps, as well as the corresponding ground-truth novel-view videos and normal maps.
Our method supports both single and multi-image inputs. Therefore, in the demonstration videos, the ``reference image(s)'' column displays a masked video, where the unmasked frames indicate the reference images. Figures \ref{fig:video_1}, \ref{fig:video_2}, and \ref{fig:video_3} show examples with only a single-view input, so only the first frame is unmasked. Figure \ref{fig:video_4} corresponds to a two-view input, hence two frames are unmasked.

The visual results demonstrate that our OrbitNVS achieves geometry- and appearance-consistent novel view synthesis, exhibiting a particularly strong capability for inferring unseen areas.

\section{Ablation Study of the VLM}\label{supsec:abla_text}

OrbitNVS incorporates text as an additional control condition. During training, the text condition is generated by a VLM that captions the full videos, while during inference, the VLM produces captions only from the reference image(s).
To investigate the role of the VLM, we conducted an ablation study on the VLM, with results shown in Tables \ref{tab:ablation_vlm_gso} and \ref{tab:ablation_vlm_omniobject}.
Specifically, in the ``w/o VLM'' setting, we replace the VLM-generated caption from the reference image with an manually-written prompt: ``\textit{Shooting around the object showcases its appearance from different perspectives.}''

The results indicate that inference without VLM captions yields comparable or marginally better performance than using them. This implies that the model's capability for unseen area synthesis is attributable to the strong visual priors in our video model, not the VLM. The VLM's captions mostly describe seen content and provide limited cues for unseen views. 
Another factor could be the limited capability of the Qwen2.5-VL-7B model \cite{bai2025qwen2}, where inaccurate captions might lead to suboptimal performance. Employing a more powerful VLM could be a promising direction for future work.


\begin{table}[h]
\tabcolsep=0.8mm
\centering
\begin{tabular}{lccc}
\toprule
Method & 0$^{\circ}$ & 30$^{\circ}$ & 60$^{\circ}$ \\
\midrule
\textbf{OrbitNVS} & 23.7 0.89 0.11 & 21.9 0.87 0.13 & 21.0 0.86 0.15  \\
w/o VLM & 24.0 0.90 0.10 & 21.9 0.87 0.12 & 21.0 0.87 0.14  \\
\bottomrule
\end{tabular}
\caption{Ablation Study of the VLM using the GSO Dataset.}
\label{tab:ablation_vlm_gso}
\vspace{-3mm}
\end{table}

\begin{table}[h]
\tabcolsep=0.8mm
\centering
\begin{tabular}{lccc}
\toprule
Method & 0$^{\circ}$ & 30$^{\circ}$ & 60$^{\circ}$ \\
\midrule
\textbf{OrbitNVS} & 21.0 0.87 0.14 & 20.4 0.86 0.14 & 19.9 0.85 0.16 \\
w/o VLM & 21.0 0.87 0.13 & 20.6 0.86 0.14 & 19.8 0.85 0.15 \\
\bottomrule
\end{tabular}
\caption{Ablation of the VLM on the OmniObject3D Dataset.}
\label{tab:ablation_vlm_omniobject}
\vspace{-3mm}
\end{table}

\section{Evaluation of Normal Map Generation}\label{supsec:normal}



OrbitNVS introduces a normal map generation branch that interacts with the main video synthesis branch to produce finer geometric details and improved consistency. We evaluate the quality of the generated normal maps both qualitatively and quantitatively. Qualitative results, provided in the supplementary videos, demonstrate that OrbitNVS generates high-quality normal maps. Quantitative results summarized in Table \ref{tab:single_view}, using metrics adopted from prior work \cite{ye2024stablenormal,fouhey2013data}, further support this observation. For instance, on the GSO dataset, the mean angular error between the predicted and ground-truth normals is less than 16°, with over 85\% of pixels exhibiting an angular error below 30$^{\circ}$.

\begin{table}[h]
\tabcolsep=1.1mm
\centering
\begin{tabular}{lcccc}
\toprule
Dataset & Mean $\downarrow$ & $<$11.25$^{\circ}$ $\uparrow$ & $<$22.5$^{\circ}$ $\uparrow$ & $<$30$^{\circ}$ $\uparrow$ \\ \midrule
\textbf{GSO} & 15.91 & 59.18 & 79.13 & 85.48\\
\textbf{OmniObject3D} & 18.72 & 48.41 & 73.78 & 82.16\\
\bottomrule
\end{tabular}
\caption{
Quantitative evaluation of normal map quality generated by OrbitNVS on both datasets. ‘Mean’ indicates the mean angular error between the predicted and ground-truth normals, and ‘$< x^{\circ}$’ reports the percentage of angles with an error less than $x^{\circ}$.
}
\label{tab:single_view}
\vspace{-3mm}
\end{table}


\end{document}